\documentclass[journal]{IEEEtran}
\def\@IEEEclspkgerror{\ClassError{IEEEtran}}
\usepackage{enumitem}
\usepackage{cite}
\usepackage{graphicx}
\usepackage{algorithm}
\usepackage{algorithmic}
\usepackage{amsmath}
\usepackage{amssymb}
\usepackage{amsthm}

\usepackage{soul}
\usepackage[dvipsnames]{xcolor}
\usepackage{hyperref}
\usepackage[dvipsnames]{xcolor}

\makeatletter
\newcounter{parenttheorem}

\makeatother

\def\BibTeX{{\rm B\kern-.05em{\sc i\kern-.025em b}\kern-.08em
    T\kern-.1667em\lower.7ex\hbox{E}\kern-.125emX}}
\usepackage{lineno}
\begin{document}

\title{ProNet: Adaptive Process Noise Estimation for INS/DVL Fusion}
\author{Barak Or and Itzik Klein
%\thanks{Submitted: May 2022. Accepted: July 2022.}
\thanks{Barak Or and Itzik Klein are with the Hatter Department of Marine Technologies, Charney School of Marine Science, University of Haifa, Haifa, 3498838, Israel (e-mail: barakorr@gmail.com, kitzik@univ@haifa.ac.il).}}

\markboth{ProNet: Adaptive Process Noise Estimation for INS/DVL Fusion / Or and Klein}%
{Or and Klein: ProNet: Adaptive Process Noise Estimation for INS/DVL Fusion}
\maketitle

\begin{abstract}
Inertial  and Doppler velocity log sensors are commonly used  to provide the navigation solution for autonomous underwater vehicles (AUV). To this end, a nonlinear filter is adopted for the fusion task. 
The filter's process noise covariance matrix is critical for filter accuracy and robustness. While this matrix varies over time during the AUV mission, the filter  assumes a constant matrix. Several models and learning approaches in the literature suggest tuning the process noise covariance during operation. In this work, we propose ProNet, a hybrid, adaptive  process, noise estimation approach for a velocity-aided navigation filter. ProNet requires only the inertial sensor reading to regress the process noise covariance. Once  learned, it is fed into the model-based navigation filter, resulting in a hybrid filter. Simulation results show the benefits of our approach compared to other models and learning adaptive approaches.  
\end{abstract}


\section{Introduction}\label{sec:introduction}
Autonomous underwater vehicles (AUVs) commonly employ an inertial navigation system (INS) and a Doppler velocity log (DVL) to estimate the AUV position, velocity, and orientation~\cite{sahoo2019advancements,9356608,karaboga2013adaptive,tal2017inertial,elhaki2020robust,eliav2018ins}. The extended Kalman filter (EKF), with an underlying assumption of constant process noise, is usually adopted for such fusion \cite{bar2004estimation,farrell2008aided}.\\
It is shown in the literature that adaptive tuning of the process noise covariance can significantly improve the filter performance.  One of the leading approaches is known as the innovation process filter \cite{mehra1970identification}, where a measure of the new information is calculated at every iteration, leading to an update of the process noise covariance matrix. Still, as recently addressed in \cite{zhang2020identification}, the question of the optimal approach to tune the process noise covariance matrix is considered open. \\
Besides the expected use of model-based approaches, learning approaches have been recently designed and integrated into AUV navigation algorithms to improve their performance \cite{klein2022}. For example, \cite{cohen2022} proposes an end-to-end network to improve the DVL velocity estimation accuracy. To cope with situations of missing beams, \cite{yona2021compensating,nadav2022} suggest a DNN to regress the missing beams. In \cite{Yinging2020}, the performance of INS/DVL fusion using a tightly coupled approach in situations of partial DVL beam measurement improved when using learning virtual beam-aided solutions. \\
In our previous works, we implemented learning approaches in the model-based Kalman filter in a hybrid fashion. Firstly, in \cite{or2022learning}, recurrent neural networks were employed to learn the vehicle’s geometrical and kinematic features to regress the process noise covariance in a linear Kalman filter framework. Later, in \cite{or2022hybrid}, we derived a hybrid learning framework to regress the process noise INS/GNSS fusion within the extended Kalman filter. Recently, we proposed a learning-based adaptive velocity-aided navigation filter \cite{OrKleinOceans}. To that end, handcrafted features were generated and used to tune the momentary system noise covariance matrix. \\
In this paper, we leverage our previous work and propose ProNet: a hybrid learning adaptive velocity-aided navigation filter using a detrend network model to regress the process noise covariance.  Once the process noise covariance is learned, it is used in the well-established, model-based, extended Kalman filter.
To validate our proposed approach, INS/DVL fusion during stimulative AUV trajectories is used. \\
The rest of the paper is organized as follows: Section II gives the model-based adaptive noise covariance approach, and Section III presents our ProNet approach. Section IV presents the analysis and results, and Section V shows the conclusions of this research.
\section{Model-based Adaptive Noise Covariance}
\label{sec:ModelAdaptive}
In its basic implementation, the error-state EKF assumes  the process noise covariance ${\bf Q}_k^d$ at each epoch to be constant. Yet, as shown in the literature \cite{zhang2020identification} and the references therein, tuning ${\bf Q}_k^d$ online improve the filter performance. \\
The most common approach to estimate ${\bf Q}_k^d$ in an adaptive error-state EKF framework was suggested in \cite{mehra1970identification, zhang2020identification}, and is based on the innovation matrix for a window of size $\xi$:
 \begin{equation} \label{eq:C}
 {{\bf{C}}_k} \buildrel \Delta \over = \frac{1}{\xi }\sum\limits_{i = k - \xi  + 1}^k {{{\bf{\nu }}_i}{{\bf{\nu }}_i}^T},
 \end{equation}
 where ${{\bf{C}}_k}$ is the innovation matrix, the innovation vector is defined by
 \begin{equation}
 {{\bf{\nu }}_k} \buildrel \Delta \over = {{\bf{v}}_{DVL,k}} - {\bf{H}} {{\bf{\hat x}}_k^ - },
 \end{equation}
$\bf{v}_{DVL,k}$ is the measured DVL velocity, $\bf{H}$ is the DVL measurement matrix, and $\bf{\hat x}_k^ -$ is the state vector.  
The innovation matrix, \eqref{eq:C}, together with the Kalman gain, ${\bf K}_k$, are used to adapt ${\bf Q}_k$, by
 \begin{equation}
 {\bf{\hat Q}}_k = {{\bf{K}}_k}{{\bf{C}}_k}{\bf{K}}_k^T.
 \end{equation}
\begin{figure*}[ht]
\centering
{\includegraphics[width=0.8\textwidth]{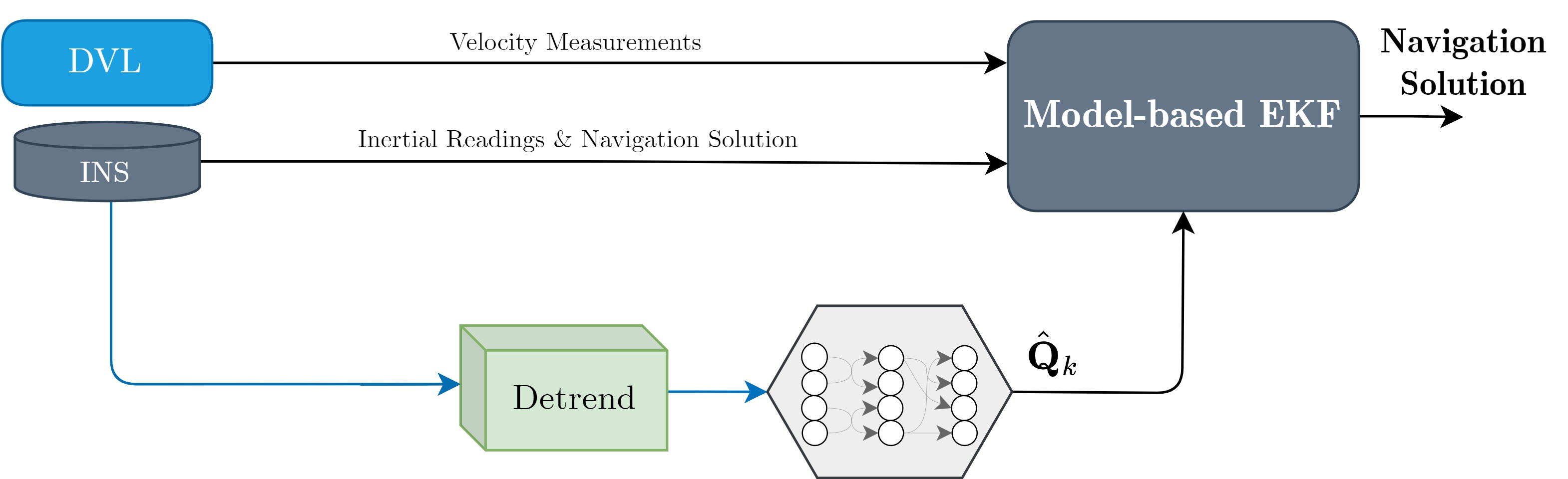}}
\caption{Boosting the hybrid adaptive navigation filter in the INS/DVL fusion with detrend transformation network for online tuning of the process noise covariance matrix.}
\label{fig:diag}
\end{figure*}
%
% %
\section{Adaptive Process Noise Estimation} 
\label{sec:Hybrid}
Leveraging from our hybrid learning adaptive navigation filter approaches \cite{or2022hybrid, OrKleinOceans}, in this research, we employ a detrend transformation on the inertial readings, prior to their use in the regression network, as a tool to improve the accuracy of the process noise covariance matrix estimation in INS/DVL fusion scenarios. Our proposed framework is shown in Figure~\ref{fig:diag}. 

\subsection{Learning the Process Noise Covariance}
% \label{sec:DataDriven}
Following \cite{or2022hybrid, OrKleinOceans} we assume that the filter continuous process noise covariance matrix at time $k$, ${\bf Q}_k^c$, is a diagonal matrix structure: 
\begin{equation}
\begin{array}{l}
{\bf{Q}}_k^c = \\
diag{\left\{ {q_f^{x*},q_f^{y*},q_f^{z*},q_\omega ^{x*},q_\omega ^{y*},q_\omega ^{z*},\varepsilon{{\bf{1}}_{1 \times 6}}} \right\}_k} \in{\mathbb{R}}{^{12 \times 12}},
\end{array}
\end{equation}
where the variance for each of the accelerometers is given by ${q_f^{i*}}$ and for the gyroscope by ${q_\omega ^{i*}}$, where  $ i \in x,y,z$. The biases are modeled by random walk processes, and their variances are set to $\varepsilon=0.001$. Our goal is to find $q_k^*$ (for all six axes) such that the speed error is minimized.  
%Formally, we search for a model to relate an instance space, ${\cal X}$, and a label space, ${\cal Y}$. We assume that there exists a target function, $\cal F$, such that ${\cal Y} = {\cal F}\left( {\cal X} \right)$. \\
%
To that end, a general one-dimensional series of length $N$ of any of the inertial sensor readings in a single axis is defined by 
\begin{equation}
{{\cal S}_k} = \left\{ {{{\bf{s}}_i}} \right\}_{i = k - N}^{k -1}.
\end{equation}
Thus, we seek ${\cal F}$, given a finite set of labeled examples (inertial sensor readings) and corresponding process noise variance values  
\begin{equation}\label{eq:sets}
\left\{ {{{\cal S}_k},q_k^*} \right\}_{k = 1}^M
\end{equation}
where $M$ is the number of examples. In this supervised learning approach, we aim to find a function ${\tilde {\cal F}}$ that best estimates ${\cal F}$. To this end, for the training process, a loss function is defined to quantify the quality of ${\tilde {\cal F}}$ with respect to ${\cal F}$. The loss function is given by
\begin{equation}\label{eq:l1}
{\cal L}\left( {{\cal Y},\hat {\cal Y}\left( {\cal X} \right)} \right) \buildrel \Delta \over = \frac{1}{M}\sum\limits_{m = 1}^M {l{{\left( {y,\hat y} \right)}_m}},
\end{equation}
where $m$ is the example index.
%Minimizing $\cal L$ in a training/test procedure leads to the target function. 
In the problem at hand, the loss function in  \eqref{eq:l1} is given by
\begin{equation}
{l_m} \buildrel \Delta \over = {\left( {q_m^* - {\hat q}_m} \right)}^2,
\end{equation}
where ${\hat q}_m$ is the estimated term obtained by the learning model during the training process.
\subsection{Network Architecture} \label{sec:Architecture}
\begin{figure*}[h!]
\centering
{\includegraphics[width=1\textwidth]{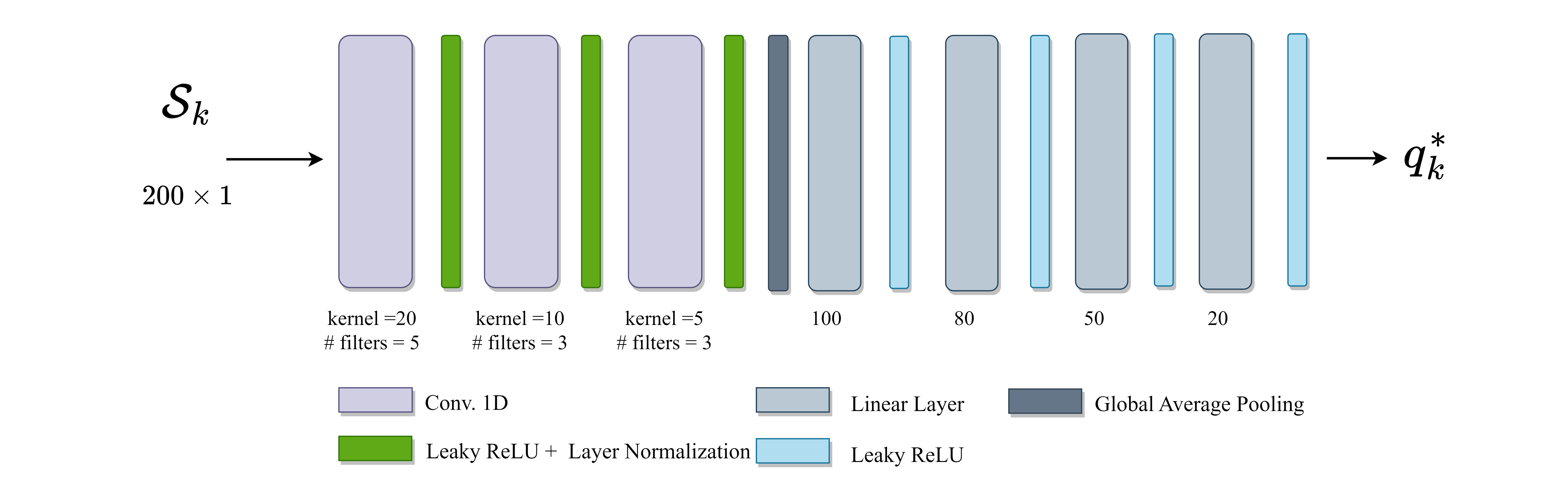}}
\caption{Baseline architecture for a single inertial sensor.}
\label{fig:baseline}
\end{figure*}

\begin{figure*}[h!]
 \centering
 {\includegraphics[width=1\textwidth]{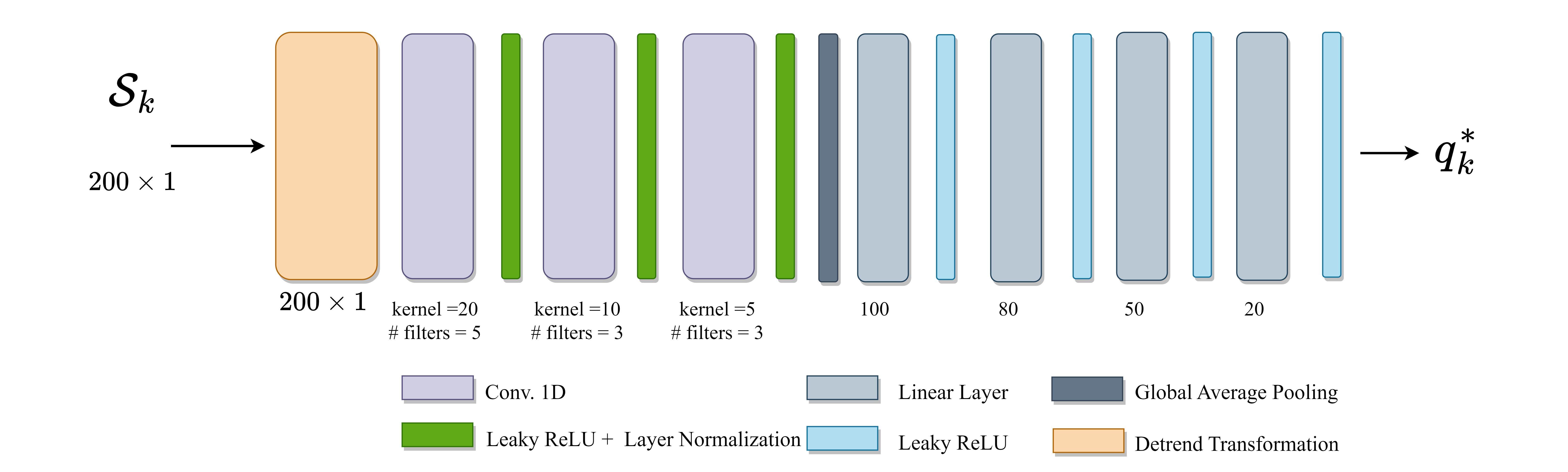}}
 \caption{Detrend-based architecture for a single inertial sensor.}
 \label{fig:net_d}
 \end{figure*}

Following our previous work in \cite{or2022hybrid}, a baseline network architecture with the following layers is defined:
\begin{enumerate}
\item {\bf Linear layer}: A layer applies a linear transformation to the incoming data from the previous layer. 
\item {\bf Conv1D layer}: A convolutional, one-dimensional (Conv1D) layer creates a convolution kernel that is convolved with the layer input over a single spatial dimension to produce a vector of outputs.
\item {\bf Global average pooling layer}: Pooling layers perform down-sampling of the feature map. They calculate the average for each input channel and flatten the data so it passes from Conv1D through the linear layers. 
\item {\bf Leaky ReLU}: The leaky rectified linear unit \cite{maas2013rectifier} is a nonlinear activation function that obtains a value $\alpha$ with the following  output:
 \begin{equation}
 f\left( \alpha  \right) = \left\{ {\begin{array}{*{20}{c}}
 \alpha &{\alpha  > 0}\\
 {0.01\alpha }&{\alpha  \le 0}
 \end{array}}. \right.
 \end{equation}
 \item {\bf Layer normalization}: Batch normalization, and in particular layer normalization, is proposed to reduce an undesirable covariate shift. The layer normalization is added after every Conv1D layer (together with the leaky ReLU layer).
\end{enumerate}
Our baseline network architecture is presented in Figure~\ref{fig:baseline}. The IMU readings are inserted into three Conv1D layers, followed by four linear layers and a global average pooling layer. Leaky ReLU activation functions are used after every layer to add nonlinearity and achieve a better generalization capability.
In addition, we propose to use the detrend transformation on the inertial readings prior to the baseline network, as illustrated in Figure~\ref{fig:net_d}. A detrend calculates the trend of the sequence and removes it from the sequence. Then, it keeps only the differences in values from the trend, allowing easier identification of cyclical and other patterns. The detrend transformation is defined by
\begin{equation}
{f_1} = detrend\left( {\cal S} \right)
\end{equation}
where ${\cal S}$ is the sequence of an inertial sensor measurement. 
\subsection{Dataset Generation} \label{sec:dataset}
\begin{figure}[ht]
\centering
{\includegraphics[width=0.4\textwidth]{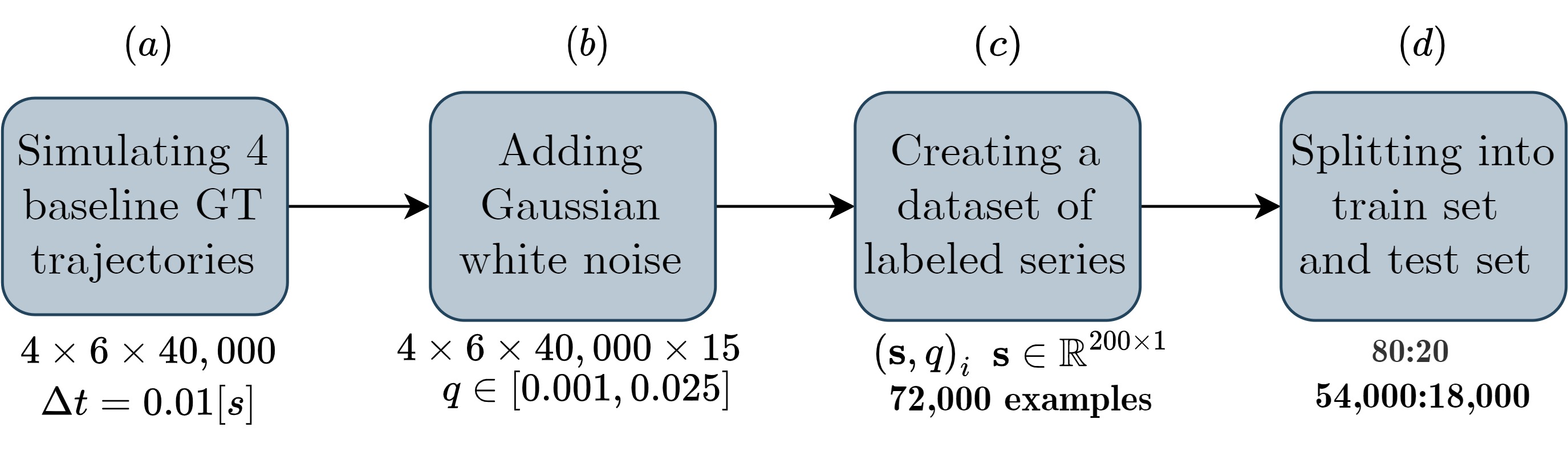}}
\caption{Dataset generation for the training phase \cite{OrKleinOceans}.}
\label{[Fig1}
\end{figure}

A stimulative-based dataset is generated, leveraging our simulation designed in \cite{or2022hybrid, OrKleinOceans}. To generate the dataset, four different baseline trajectories are constructed and divided into training and testing sets as illustrated in Figure~\ref{[Fig1}. The richness of the baseline trajectories allows the creation of a model that is able to cope with unseen trajectories. Each baseline trajectory is created by  assuming perfect inertial readings for a period of $400$s with a sampling rate of $100$Hz, resulting in a sequence of $6\times40,000$ samples for each baseline trajectory.
To emulate real-world varying conditions, to each of the six inertial channels we add an additive zero-mean white Gaussian noise with variance in the range of $q\in[0.001,0.05]$, with 15 different values inside this interval.
Thus, each baseline trajectory has $15$ series of $6\times40,000$ noisy inertial samples. %The justification to include a simple noise model lies in the momentary IMU measurement noise covariance sequence as a short time window is considered and thus allows characterizing the noise with its variance only.
Next, a series of length $N$ is chosen, and a corresponding labeled database is generated. Lastly, we choose $N=200$ samples and create batches corresponding to two seconds each with a total of $6\times200\times 15$ per baseline trajectory. Then, these batches are randomly divided into training and testing datasets, ensuring that all baseline trajectories are included both in the train and the test sets with a ratio of 80:20. In total, there are 72,000 examples; 57,600 in the train set and 14,400 in the test set. 
% %
% \begin{figure}[ht]
% \centering
% {\includegraphics[width=0.35\textwidth]{four_trajectories.jpg}}
% \caption{Our four baseline trajectories used to create the dataset. }
% \label{fourbaseline}
% \end{figure}
\subsection{ProNet Algorithm} \label{sec:HybridScheme}
The ProNet algorithm is presented as Algorithm 1. It describes the INS/DVL fusion, allowing process noise covariance learning in real-time conditions. Note that the inertial readings are used  both in the navigation filter and in the regression network. 
\begin{algorithm}[h!]
 \caption{ProNet with detrend network architecture}
 \begin{algorithmic}[1]
 \renewcommand{\algorithmicrequire}{\textbf{Input:}}
 \renewcommand{\algorithmicensure}{\textbf{Output:}}
 \REQUIRE ${\bf \omega}_{ib},{\bf f}_b,{\bf{v}}_{Aiding},\Delta t_0, \Delta \tau,T,tuningRate$
 \ENSURE  ${\bf{v}}^n,{\bf \varepsilon}^n$
 \\ \textit{Initialization} : ${\bf{v}}^n_0,{\bf{\varepsilon}}^n_0$
 \\ \textit{LOOP Process}
   \FOR {$t = 0$ to $T$}
   \STATE Get inertial readings ${\bf \omega}_{ib},{\bf f}_b$ 
  \STATE Solve navigation equations of motion
  \IF {$(mod (t,\Delta \tau)$=0)}
  \STATE Get DVL measurement ${\bf{v}}_{DVL}$ 
 \STATE  Apply the EFK navigation filter
  \ENDIF
  \STATE Predict ${\bf Q}_{k+1}^c$ based on ProNet model
  \IF {$mod(t,tuningRate)=0$}
\STATE ${\bf{Q}}_{k + 1}^c \leftarrow {\bf{\hat Q}}_{k + 1}^c\left( {{{\cal S}_k}} \right)$
\ENDIF
  \ENDFOR
 \end{algorithmic}
 \end{algorithm}
\section{Analysis and Results} \label{sec:Results}
\begin{figure}[ht]
\centering
{\includegraphics[width=0.48\textwidth]{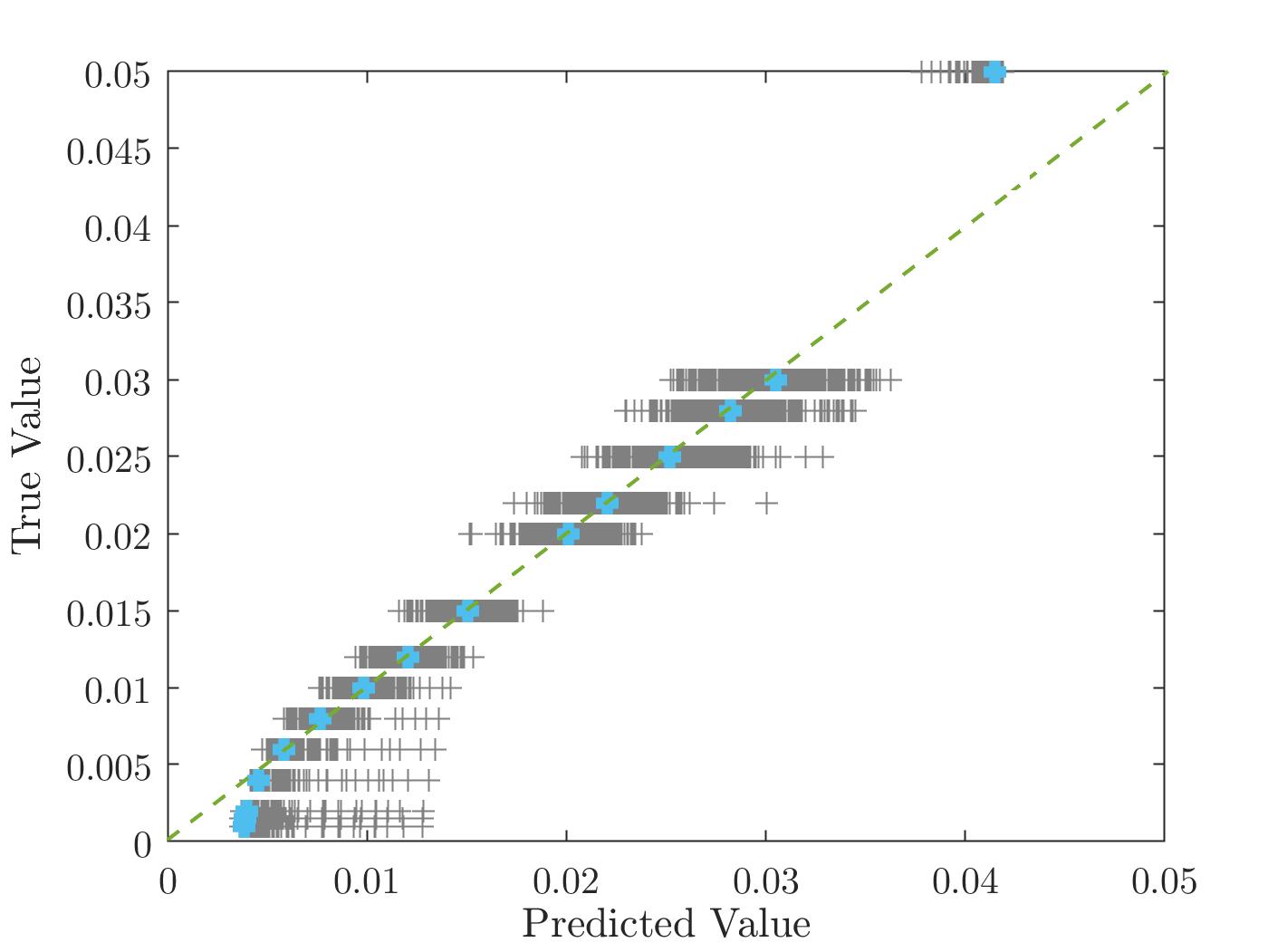}}
\caption{Detrend-network: ground-truth values versus predicted network values on the test dataset.}
\label{[Fig8}
\end{figure}

Two metrics are employed to evaluate the speed accuracy of the proposed approach:
\begin{itemize}
    \item Speed root mean squared error ($2^{nd}$ norm) for all three axes:
\begin{equation}
SRMSE = \sqrt {\frac{1}{y}\sum\limits_{k = 1}^{y}  {\sum\limits_{j \in \left\{ {N,E,D} \right\}}^{} {\delta {{\hat v}_{jk}}^2} } } .
\end{equation}
    \item Speed mean absolute error for all three axes:
\begin{equation}
SMAE = \frac{1}{y}\sum\limits_{k = 1}^{y} {\sum\limits_{j \in \left\{ {N,E,D} \right\}}^{} {\left| {\delta {{\hat v}_{jk}}} \right|}},
\end{equation}
\end{itemize}
where $y$ is the number of samples, $k$ is a running index for the samples, $j$ is the running index for the velocity component in the North, East, and Down (NED) directions, and $\delta {\hat v}_{jk}$ is the velocity error term. \\
Our proposed detrend network is compared to the baseline approach, to our previous handcrafted feature ensemble model~\cite{OrKleinOceans}, to the model-based adaptive innovation approach (1)-(3), and to the constant process noise covariance with two different values. To that end, an AUV trajectory traveling in straight lines with left/right turning and diving is generated.  All the simulation parameters are presented in Table~\ref{table2}. All cases of the model-based EKF, including two cases with a constant process noise covariance matrix, are examined by performing 100 Monte Carlo (MC) simulations.
\begin{table}[ht!]
\label{table2}
\caption {INS/DVL navigation filter parameters.} \label{tab:title} 
\setlength{\tabcolsep}{3pt}
\begin{center}
\begin{tabular}{|p{90pt}|p{60pt}|p{75pt}|}\hline
\bf{Description} & \bf{Symbol} & \bf{Value} \\
\hline
DVL noise (var)          & $R_{11},R_{22},R_{33}$      &$0.01{[m/s]^2}$   \\ 
DVL step size &$\Delta \tau$ & $1 [s]$   \\ 
Accelerometer noise (var) & $Q_{11},Q_{22},Q_{33}$   & $0.01^2[m/s^2]^2$      \\ Gyroscope noise (var) & $Q_{44},Q_{55},Q_{66}$   & $0.001^2[rad/s]^2$ \\ Simulation duration   & $T$   &$330 [s]$    \\
Initial velocity   & ${\bf{v}}^n_0$   &$[1, 0, 0]^T [m/s]$    \\
Initial position   & ${\bf{p}}^n_0$   & $[32^0, 34^0,-5[m]]^T $    \\
\hline
\end{tabular}
\end{center}
\end{table}
Figure~\ref{[Fig8} presents the prediction results on the test dataset for our detrend network. The red line is the desired one, representing when the ground truth value is equal to the learning predicted value. The gray points represent the performance of the test set with the trained learning model. The mean values of the test set lie near the red line (blue points) in most cases.
The results in terms of SRMSE and SMAE are presented in Table~\ref{tab:title} for the model- and hybrid-based approaches. In the first line, the process noise covariance matrix is tuned according to the actual IMU covariance values. It obtains an SRMSE of 1.082[m/s] and SMAE of 0.961[m/s]. Then, we examined a case where the true noise covariance matrix values are multiplied by 20. As expected, this erroneous model-based approach obtains worse results than the previous one with an SRMSE of 1.142 [m/s] and SMAE of 0.992[m/s]. Applying the innovation-based approach (lines three and four) leads to an improvement of 5.5\% in the SMAE. \\
Our ProNet approaches show an improvement compared to all the model-based methods. The baseline network obtains an SRMSE of 1.055[m/s] and SMAE of 0.928[m/s], improvements of 2.5\% and 3.4\%, respectively, over the first (constant) case. Our HCF-Ensemble hybrid model approach achieves an SRMSE of 0.98[m/s] and SMAE of 0.866[m/s], showing improvements of 10.4\% and 9.8\%, respectively, over the first (constant) case. It also improves the baseline  approach by more than $9\%$. The detrend network obtains the lowest SMAE value of 0.831[m/s].
\begin{table}[h!]
\caption {Experiment results for several model and learning approaches.} \label{tab:title}
\begin{center}
\begin{tabular}{ |c|c|c| } 
\hline
$\bf Approach$ &${\bf  SRMSE} [m/s]$ &${\bf  SMAE} [m/s]$\\
\hline
Constant $q_f^i=0.01$ $q_{\omega}^i=0.001$  & 1.082  & 0.961   \\ 
\hline
Constant $q_f^i=0.2$ $q_{\omega}^i=0.02$ &  1.142    & 0.992 \\ 
\hline
Adaptive $\xi=1$  & 1.127 & 0.983  \\ 
\hline
Adaptive $\xi=5$  & 1.085 & 0.908  \\ 
\hline
ProNet: HCF-ensemble & {\bf{0.980}}   & {0.866} \\ 
\hline
ProNet: baseline network & 1.055    & 0.928 \\ 
\hline
ProNet: detrend network & {\bf{0.980}}    & {\bf{0.831}} \\
\hline
\end{tabular}
\end{center}
\end{table}
\section{Conclusions} \label{sec:Conc}
The process noise covariance has a great influence on INS/DVL filter performance, therefore, a good choice is critical for accurate AUV navigation. In this research, a hybrid learning model was suggested in an adaptive extended Kalman filter framework. A detrend network, where the inertial readings are passed via a detrend transformation, was suggested to adaptively regress the process noise covariance. This network was plugged into
the model-based EKF with an error-state implementation, resulting in a hybrid learning adaptive navigation filter.\\
Our network was trained on a simulated dataset consisting of four different baseline trajectories. To validate the performance of the proposed approach, it was compared to  model-based approaches with constant and adaptive process noise covariance and both of our previous approaches; namely, the baseline network and handcrafted feature ensemble approaches.  
Simulation results on the test dataset show that our  detrend network obtained an SRMSE of 0.98[m/s], an improvement of about 10\% over a constant noise covariance matrix, and a minimum SMAE of 0.831[m/s]. 

Although demonstrated here for AUV navigation with INS/DVL fusion, the proposed approach can be elaborated upon for any external sensor aiding  the INS and for any type of platform.

\bibliographystyle{IEEEtran}
\bibliography{IEEEfull}

\begin{thebibliography}{10}
\providecommand{\url}[1]{#1}
\csname url@samestyle\endcsname
\providecommand{\newblock}{\relax}
\providecommand{\bibinfo}[2]{#2}
\providecommand{\BIBentrySTDinterwordspacing}{\spaceskip=0pt\relax}
\providecommand{\BIBentryALTinterwordstretchfactor}{4}
\providecommand{\BIBentryALTinterwordspacing}{\spaceskip=\fontdimen2\font plus
\BIBentryALTinterwordstretchfactor\fontdimen3\font minus
  \fontdimen4\font\relax}
\providecommand{\BIBforeignlanguage}[2]{{%
\expandafter\ifx\csname l@#1\endcsname\relax
\typeout{** WARNING: IEEEtran.bst: No hyphenation pattern has been}%
\typeout{** loaded for the language `#1'. Using the pattern for}%
\typeout{** the default language instead.}%
\else
\language=\csname l@#1\endcsname
\fi
#2}}
\providecommand{\BIBdecl}{\relax}
\BIBdecl

\bibitem{sahoo2019advancements}
A.~Sahoo, S.~K. Dwivedy, and P.~Robi, ``Advancements in the field of autonomous
  underwater vehicle,'' \emph{Ocean Engineering}, vol. 181, pp. 145--160, 2019.

\bibitem{9356608}
Y.~Yang, Y.~Xiao, and T.~Li, ``A survey of autonomous underwater vehicle
  formation: Performance, formation control, and communication capability,''
  \emph{IEEE Communications Surveys and Tutorials}, vol.~23, no.~2, pp.
  815--841, 2021.

\bibitem{karaboga2013adaptive}
N.~Karaboga and F.~Latifoglu, ``Adaptive filtering noisy transcranial {D}oppler
  signal by using artificial bee colony algorithm,'' \emph{Engineering
  Applications of Artificial Intelligence}, vol.~26, no.~2, pp. 677--684, 2013.

\bibitem{tal2017inertial}
A.~Tal, I.~Klein, and R.~Katz, ``Inertial navigation system/{D}oppler velocity
  log ({INS/DVL}) fusion with partial {DVL} measurements,'' \emph{Sensors},
  vol.~17, no.~2, p. 415, 2017.

\bibitem{elhaki2020robust}
O.~Elhaki and K.~Shojaei, ``A robust neural network approximation-based
  prescribed performance output-feedback controller for autonomous underwater
  vehicles with actuators saturation,'' \emph{Engineering Applications of
  Artificial Intelligence}, vol.~88, p. 103382, 2020.

\bibitem{eliav2018ins}
R.~Eliav and I.~Klein, ``{INS}/partial {DVL} measurements fusion with
  correlated process and measurement noise,'' \emph{Multidisciplinary Digital
  Publishing Institute Proceedings}, vol.~4, no.~1, p.~34, 2018.

\bibitem{bar2004estimation}
Y.~Bar-Shalom, X.~R. Li, and T.~Kirubarajan, \emph{Estimation with applications
  to tracking and navigation: theory algorithms and software}.\hskip 1em plus
  0.5em minus 0.4em\relax John Wiley \& Sons, 2004.

\bibitem{farrell2008aided}
J.~Farrell, \emph{Aided navigation: GPS with high rate sensors}.\hskip 1em plus
  0.5em minus 0.4em\relax McGraw-Hill, Inc., 2008, pp: 393-395.

\bibitem{mehra1970identification}
R.~Mehra, ``On the identification of variances and adaptive {K}alman
  filtering,'' \emph{IEEE Transactions on {A}utomatic {C}ontrol}, vol.~15,
  no.~2, pp. 175--184, 1970.

\bibitem{zhang2020identification}
L.~Zhang, D.~Sidoti, A.~Bienkowski, K.~R. Pattipati, Y.~Bar-Shalom, and D.~L.
  Kleinman, ``On the identification of noise covariances and adaptive {K}alman
  filtering: A new look at a 50 year-old problem,'' \emph{IEEE Access}, vol.~8,
  pp. 59\,362--59\,388, 2020.

\bibitem{klein2022}
I.~Klein, ``Data-driven meets navigation: Concepts, models, and experimental
  validation,'' in \emph{2022 DGON Inertial Sensors and Systems (ISS)}, 2022,
  pp. 1--21.

\bibitem{cohen2022}
N.~Cohen and I.~Klein, ``Beamsnet: {A} data-driven {A}pproach {E}nhancing
  {D}oppler {V}elocity {L}og {M}easurements for {A}utonomous {U}nderwater
  {V}ehicle {N}avigation,'' \emph{Engineering Applications of Artificial
  Intelligence, Vol. 114, 105216}, 2022.

\bibitem{yona2021compensating}
M.~Yona and I.~Klein, ``Compensating for {P}artial {D}oppler {V}elocity {L}og
  {O}utages by {U}sing {D}eep-{L}earning {A}pproaches,'' in \emph{2021 IEEE
  International Symposium on Robotic and Sensors Environments (ROSE)}.\hskip
  1em plus 0.5em minus 0.4em\relax IEEE, 2021, pp. 1--5.

\bibitem{nadav2022}
N.~Cohen and I.~Klein, ``{AUV} {V}elocity {V}ector {E}stimation in {S}ituations
  of {L}imited {DVL} {B}eam {M}easurements,'' \emph{IEEE Oceans 2022, Hampton
  Roads}, 2022.

\bibitem{Yinging2020}
Y.~Yao, X.~Xu, X.~Xu, and I.~Klein, ``Virtual beam aided {SINS/DVL} tightly
  coupled integration method with partial {DVL} measurements,'' \emph{IEEE
  Transactions on Vehicular Technology}, 2022.

\bibitem{or2022learning}
B.~Or and I.~Klein, ``Learning {V}ehicle {T}rajectory {U}ncertainty,''
  \emph{arXiv preprint arXiv:2206.04409}, 2022.

\bibitem{or2022hybrid}
B.~{O}r and I.~Klein, ``A {H}ybrid {M}odel and {L}earning-{B}ased {A}daptive
  {N}avigation {F}ilter,'' \emph{IEEE Transactions on Instrumentation and
  Measurement}, pp. 1--1, 2022.

\bibitem{OrKleinOceans}
B.~Or and I.~Klein, ``A hybrid adaptive velocity aided navigation filter with
  application to {INS/DVL} fusion,'' in \emph{OCEANS 2022: Hampton Roads},
  2022, pp. 1--6.

\bibitem{maas2013rectifier}
A.~L. Maas, A.~Y. Hannun, A.~Y. Ng \emph{et~al.}, ``Rectifier nonlinearities
  improve neural network acoustic models,'' in \emph{Proc. icml}, vol.~30,
  no.~1.\hskip 1em plus 0.5em minus 0.4em\relax Citeseer, 2013, p.~3.

\end{thebibliography}

\end{document}